\documentclass{article}

\usepackage{iclr2027_conference,times}

\usepackage[utf8]{inputenc}
\usepackage[T1]{fontenc}
\usepackage{hyperref}
\hypersetup{
  hidelinks,
  pdftitle={UniCache: Task- and Type-Aware KV Cache Compression for Unified Multimodal Models},
  pdfauthor={Wanqi Yang, Yuexiao Ma, Mei Xie, Xiawu Zheng, Shiwei Liu}
}
\usepackage{url}
\usepackage{booktabs}
\usepackage{amsfonts}
\usepackage{amsmath}
\usepackage{amssymb}
\usepackage{nicefrac}
\usepackage{microtype}
\usepackage{xcolor}
\definecolor{unicachepink}{HTML}{B93F78}
\usepackage{graphicx}
\usepackage{multirow}
\usepackage{float}
\usepackage{wrapfig}
\usepackage{enumitem}
\usepackage{caption}

\setcitestyle{round,semicolon,aysep={,}}
\newcommand{\designprinciple}[2]{%
  \par\smallskip\noindent
  \colorbox{black!5}{%
    \parbox{\dimexpr\linewidth-2\fboxsep\relax}{%
      \textbf{Design Principle #1:}\enspace
      \textit{#2}%
    }%
  }\par\smallskip
}
\title{{\LARGE UniCache: Task- and Type-Aware KV Cache\\Compression for Unified Multimodal Models}}

\author{%
    \textbf{Wanqi Yang}$^{1,2,3}$\hspace{0.55em}\textbf{Yuexiao Ma}$^{4}$\hspace{0.55em}\textbf{Mei Xie}$^{5}$\hspace{0.55em}\textbf{Xiawu Zheng}$^{6}$\hspace{0.55em}\textbf{Shiwei Liu}$^{1,2,3}$\\[0.4em]
    {\normalfont\footnotesize $^{1}$Max Planck Institute for Intelligent Systems \quad $^{2}$ELLIS Institute T\"ubingen}\\
    {\normalfont\footnotesize $^{3}$T\"ubingen AI Center \quad $^{4}$Nanyang Technological University}\\
    {\normalfont\footnotesize $^{5}$Independent Researcher}\\
    {\normalfont\footnotesize $^{6}$Key Laboratory of Multimedia Trusted Perception and Efficient Computing, Xiamen University}
}

\iclrfinalcopy
\begin{document}
	
	\maketitle
	\vspace{-0.22in}
	{\centering\small
	\href{https://superone77.github.io/UniCache/}{\textcolor{unicachepink}{\textbf{Project Page}}}
	\hspace{1.5em}
	\href{https://github.com/Superone77/UniCache}{\textcolor{unicachepink}{\textbf{Code}}}\par}
	\vspace{0.14in}

\begingroup
\setlength{\intextsep}{2pt}
\setlength{\abovecaptionskip}{2pt}
\setlength{\belowcaptionskip}{6pt}
\captionsetup{font=small}
\begin{figure}[H]
    \centering
    \includegraphics[width=\linewidth]{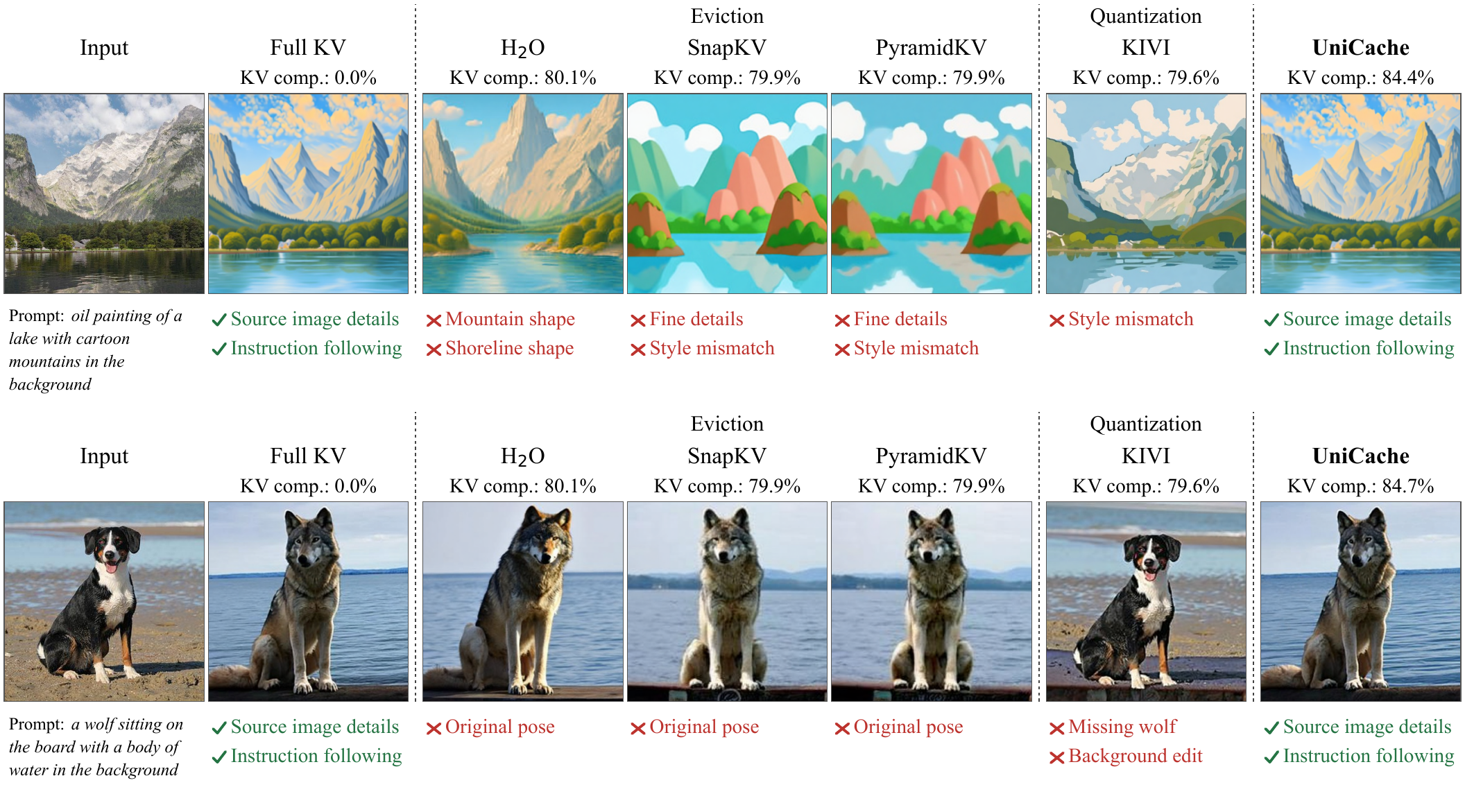}
    \caption{\textbf{UniCache versus uniform KV-cache compression in image editing.} At approximately 80\% compression, eviction tends to lose source-image details while quantization weakens instruction following. UniCache preserves both.}
    \label{fig:fig1_editing_comparison_extended}
\end{figure}
\endgroup

	\begin{abstract}
        \looseness=-1 Unified multimodal models combine understanding, generation, and editing within a single network, offering a promising foundation for versatile multimodal applications. However, growing multimodal contexts make KV cache storage and access increasingly costly. Existing KV cache compression methods are typically tailored to specific tasks and single-modality caches, while overlooking changes in cache importance across tasks and timesteps. However, in unified multimodal models, each task involves multiple KV cache types, and both their composition and dynamics differ across tasks. As a result, a single compression policy overlooks task- and type-specific requirements, leading to the loss of critical information and degraded quality across tasks. Based on these findings, we propose \textbf{UniCache}, a training-free framework for task- and type-aware KV cache compression. \textsc{UniCache} identifies the cache segments activated by each task and assigns suitable compression policies through offline calibration. It coordinates their parallel execution under a shared storage budget through attention-guided allocation and task-aware temporal scheduling. Experiments show that \textsc{UniCache} achieves \textbf{5}$\times$ KV cache compression for understanding and editing and \textbf{2.5}$\times$ for generation with negligible quality loss, while increasing throughput by up to \textbf{1.78}$\times$ in long-context settings, significantly improving the practicality of scaling unified multimodal models to longer context.
        
    \end{abstract}

	\section{Introduction}
    \begin{figure}[b!]
    \centering
    \includegraphics[width=\linewidth]{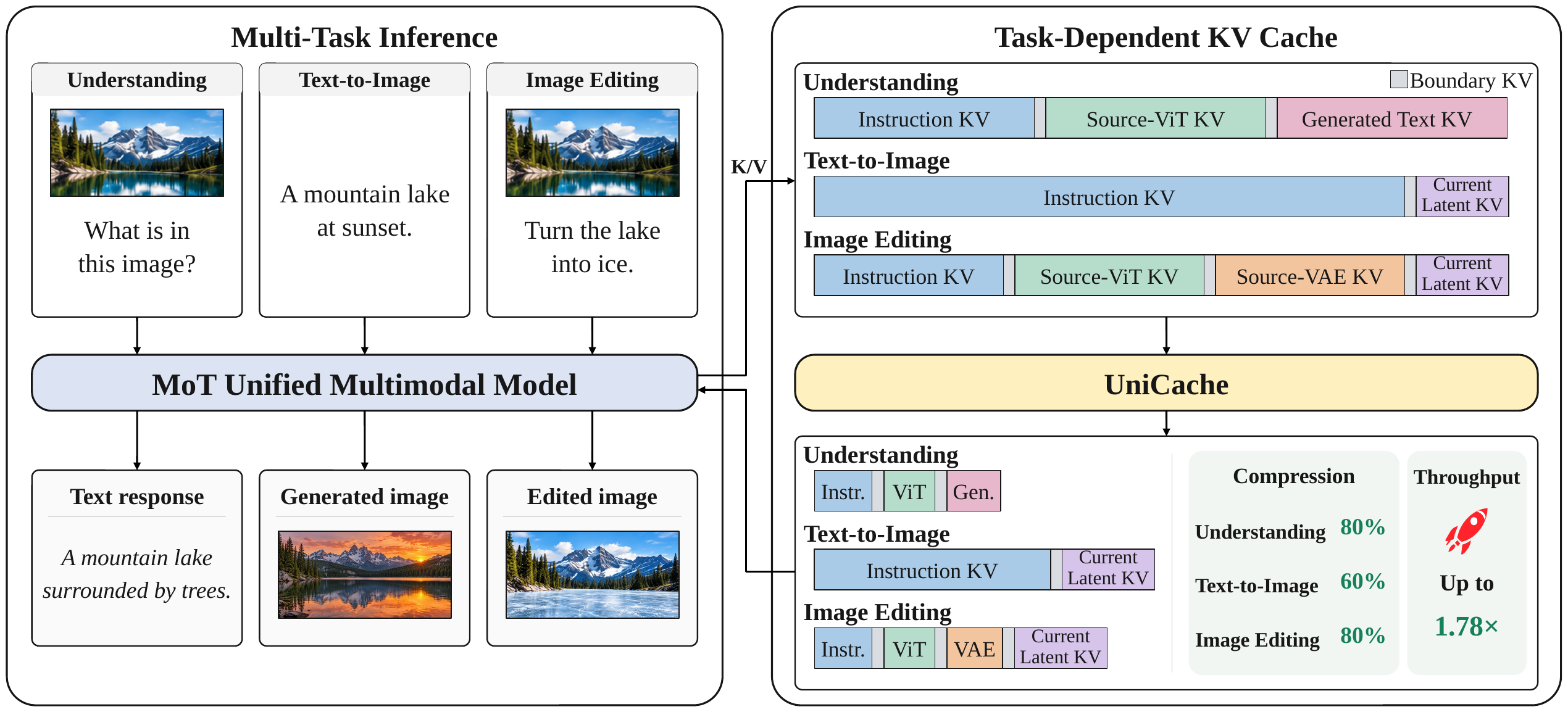}
    \caption{\textbf{Left}: Examples of deploying a Mixture-of-Transformers (MoT) unified multimodal model for multi-task inference: image understanding, text-to-image generation and image editing. \textbf{Right}: UniCache adopts task- and type-aware KV cache compression and significantly increases throughput and reduces memory overhead for KV cache.}
    \label{fig:kv_cache_type_in_task}
    \end{figure}
	\looseness=-1 Unified multimodal models integrate understanding and generation within a single network, enabling visual understanding to support image creation and editing from multimodal instructions~\citep{team2024chameleon,zhou2025transfusion,xie2025show,wu2025janus}. Recent Mixture-of-Transformers (MoT) models, exemplified by BAGEL, combine modality-specific Transformer experts through shared attention to support these capabilities~\citep{liang2024mixture,deng2025emerging,diao2026sensenova}. However, processing increasingly rich multimodal contexts requires storing and repeatedly accessing large KV caches during inference~\citep{shazeer2019fast,kwon2023efficient}. As context length grows, the resulting memory footprint and access overhead constrain inference efficiency. KV cache compression therefore offers a practical way to make these models more efficient and support longer multimodal contexts.

    \looseness=-1 Existing KV cache compression methods reduce memory costs through token eviction and quantization~\citep{xiao2024efficient,zhang2023H2O,cai2024pyramidkv,he2024zipvl, liu2024kivi,tu2025vl,yang2026alphaq}. Most are designed for a fixed inference paradigm, whereas unified models must manage semantically distinct cache types whose roles change across tasks and inference stages. This challenge is particularly evident in image editing, where the model must follow an editing instruction while preserving source-image content unrelated to the requested change. As shown in Figure~\ref{fig:fig1_editing_comparison_extended}, eviction-based methods tend to lose source-image details, while quantization preserves those details relatively well but fails to apply the requested edit. These contrasting failures show why a single global compression policy struggles to preserve quality across tasks.

	Our analysis of KV cache behavior across tasks in unified multimodal models reveals four challenges underlying these limitations:
\begin{itemize}[
    leftmargin=1.2em,
    labelsep=0.4em,
    itemsep=1pt,
    topsep=2pt,
    parsep=0pt,
    partopsep=0pt
]
\item \textbf{Task-dependent cache composition.}
Unified multimodal models contain multiple KV cache types within each task, and the types involved differ across tasks. Compression therefore cannot assume a fixed set of cache types across all task modes.

\item \textbf{Type-dependent compression preferences.}
KV cache types within the same task differ in their sensitivity to token eviction and quantization. A uniform policy may preserve the information in one KV cache type effectively while disproportionately degrading another.

\item \textbf{Unequal importance across cache types.}
Different KV cache types contribute unequally to attention, so that applying a uniform compression budget cannot account for these differences.

\item \textbf{Task- and stage-dependent cache importance.}
The relative importance of cache types and the overall importance of KV cache evolve differently across tasks and inference stages. 
\end{itemize}
\begin{figure}[t]
		\centering
		\includegraphics[width=\textwidth]{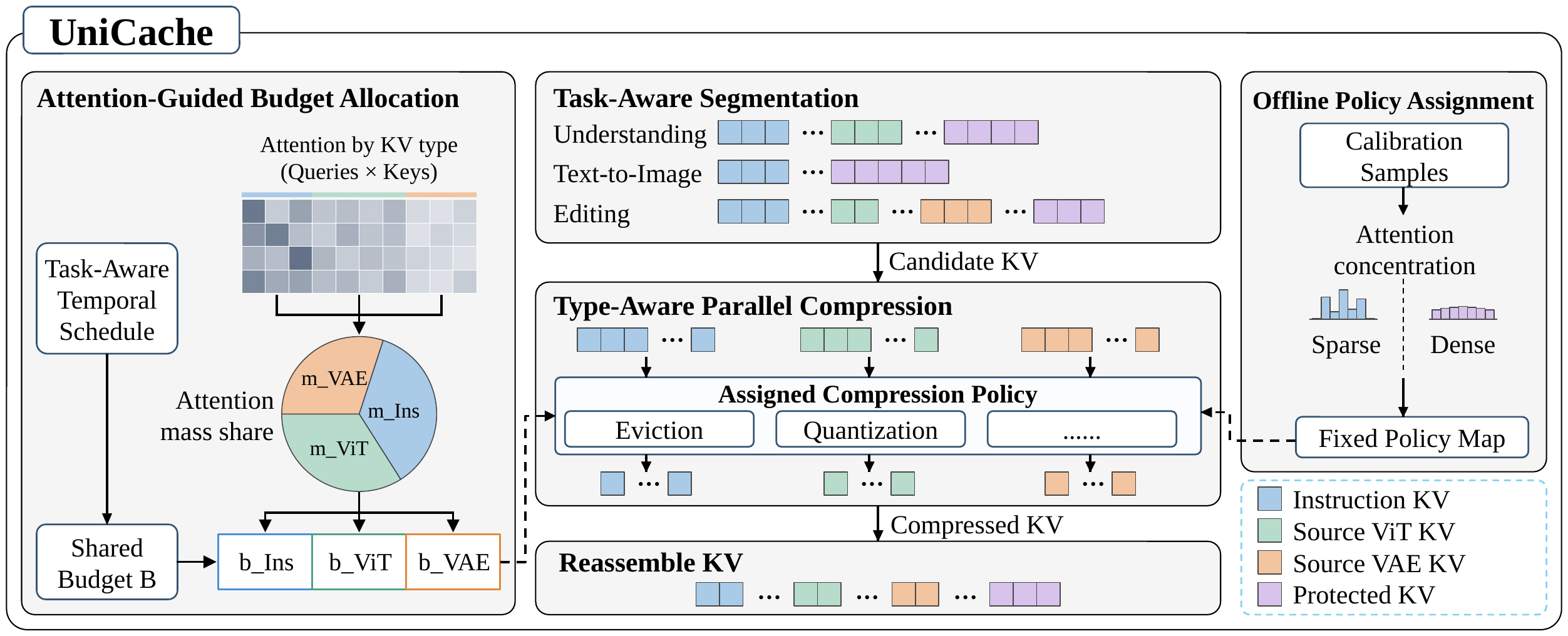}
		\caption{\textbf{Overview of UniCache.} It identifies the cache segments activated by each task, assigns suitable compression policies via offline calibration and allocates type-specific compression budgets from the corresponding attention statistics, compressing different cache types online and in parallel.}
		\label{fig:framework}
	\end{figure}

    To address these challenges, we propose \textbf{UniCache}, a training-free framework for task- and type-aware KV cache compression. UniCache identifies the KV cache types present in each task and separates compression candidates from protected segments. Through offline calibration of attention distributions, UniCache assigns suitable compression policies to different candidate types. During inference, UniCache determines each segment's storage budget through attention-guided allocation and task-aware temporal scheduling, then applies the assigned compression policies independently and in parallel to their respective segments.

    We demonstrate that, across different unified multimodal models and benchmarks, UniCache better preserves quality than uniform compression policies under comparable storage budgets. Figure~\ref{fig:fig1_editing_comparison_extended} illustrates how UniCache preserves both source-image details and instruction following. With our efficient inference implementation, UniCache further reduces KV cache memory usage and improves long-context throughput. Notably, on BAGEL, UniCache achieves $5\times$ KV cache compression for understanding and editing and $2.5\times$ for generation with negligible quality loss, while increasing throughput by up to $1.78\times$ in long-context settings.

\section{Related Work}
	\label{sec:related}
	\paragraph{Unified multimodal models and inference acceleration.}
	Unified multimodal models integrate visual understanding and visual generation within a single architecture, and recent designs pursue this integration along markedly different paradigms~\citep{team2024chameleon,zhou2025transfusion,xie2025show,wu2025janus}. BAGEL~\citep{deng2025emerging} adopts a Mixture-of-Transformers (MoT) architecture, supporting understanding, image generation, and image editing within one model. As the capability coverage of such models grows, their inference efficiency has drawn increasing attention. Hyper-Bagel~\citep{lu2025hyperbagel} accelerates both task modes within a single framework, applying speculative decoding to the autoregressive prefill--decode path used for understanding and diffusion distillation to the iterative denoising used for generation. Flash-Unified~\citep{ke2026flash} goes further by exploiting the distinct computational profiles of the two task modes, combining per-task network pruning and dynamic layer skipping with visual token pruning for understanding and diffusion-head caching for generation. Together, these methods indicate that unified models differ significantly in their execution characteristics across different tasks, and that inference optimization must therefore be mode-aware. However, the heterogeneity of the KV cache that arises across task modes and across semantically distinct KV cache types remains underexplored.
	
	\paragraph{KV cache compression.}
	Existing KV cache compression methods are designed primarily for autoregressive models, where they reduce cache overhead through token eviction or sparse access. StreamingLLM~\citep{xiao2024efficient} exploits the attention sink phenomenon by retaining a small set of critical initial tokens, while H$_2$O~\citep{zhang2023H2O} and Scissorhands~\citep{liu2023scissorhands} decide which KV entries to keep based on the accumulated importance of each token. SnapKV~\citep{li2024snapkv} and PyramidKV~\citep{cai2024pyramidkv} go further and use observed attention patterns to guide both token selection and the allocation of the cache budget across layers, and DiffKV~\citep{zhang2025diffkv} introduces differentiated memory management for keys and values. However, these designs do not jointly account for the two dimensions of heterogeneity that characterize MoT-based unified multimodal models: task mode and cache type.  UniCache fills this gap by coupling task-aware and type-aware cache management: it assigns a compression strategy per cache type and dynamically allocates a shared cache budget according to the task and the current inference stage.

\section{UniCache}
\label{sec:unicache}

In this section, we introduce  \textbf{UniCache}. We first analyze the challenges that unified multimodal models introduce to KV cache compression, then derive four design principles and explain how UniCache addresses these challenges through coordinated policy assignment and budget management. Figure~\ref{fig:framework} presents the resulting framework.

\begin{figure}[t]
	\centering
	\includegraphics[width=1\linewidth]{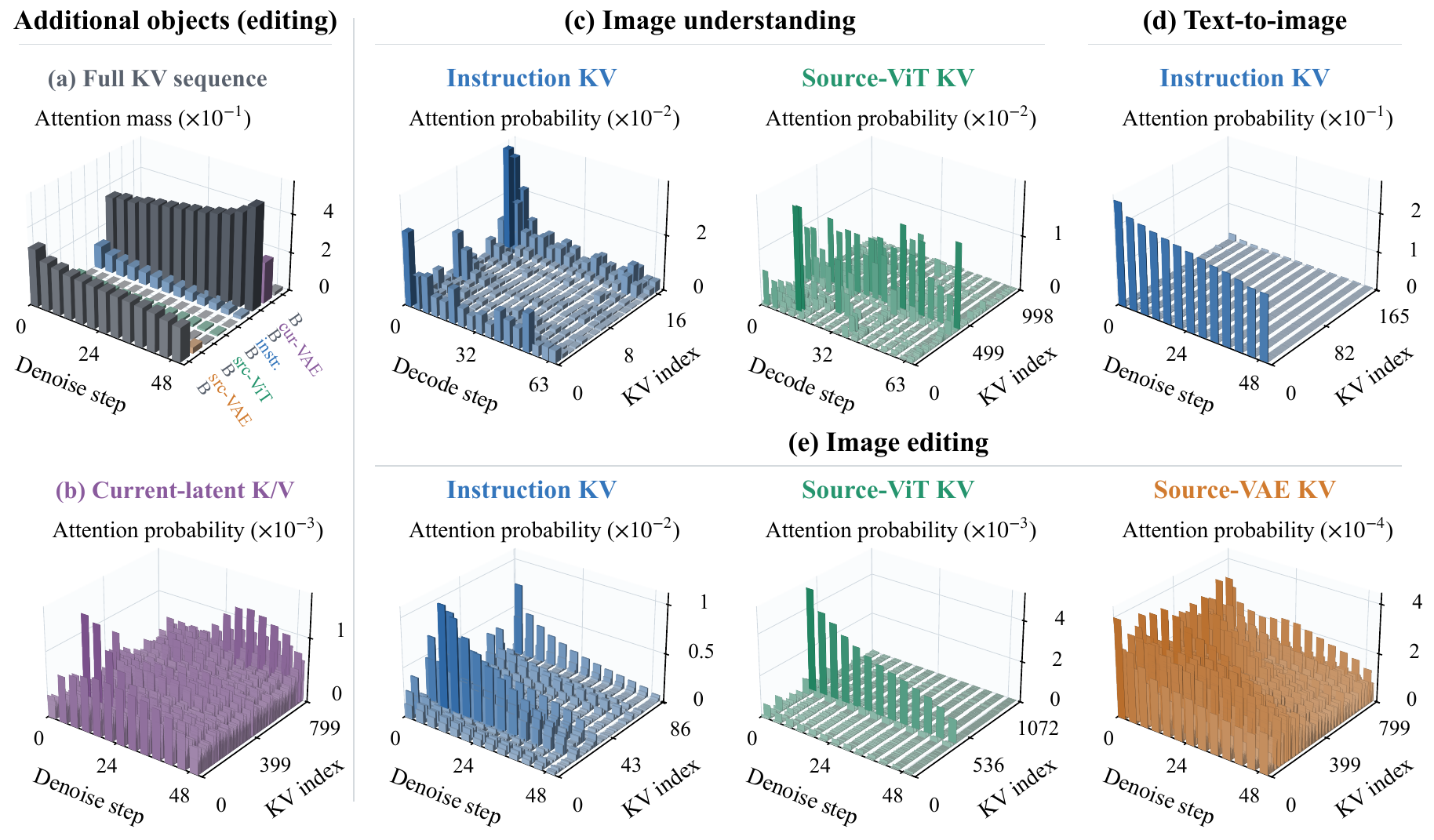}
	\caption{\textbf{Task- and type-dependent KV cache attention patterns in BAGEL at layer 8.} (a) Attention mass over the complete editing KV sequence, preserving the original segment order, where \emph{B} denotes boundary KV. (b) Current-latent K/V recomputed at each denoising step.  (c--e) Attention distributions of different cache types across image understanding, text-to-image generation, and image editing. We report attention probability averaged over heads and queries.
	}
	\label{fig:fig_task_type_cache_distributions_L08}
\end{figure}
\subsection{Task-Aware Cache Segmentation}

\looseness=-1 \paragraph{Task-dependent cache composition.} In MoT-based unified multimodal models, the information required by different tasks is represented by distinct KV cache types, as illustrated in Figure~\ref{fig:kv_cache_type_in_task}. In BAGEL, image understanding uses instruction and source-ViT KVs to represent the instruction and the source image's semantic information, respectively. Text-to-image generation conditions on instruction KV, while image editing uses instruction, source-ViT, and source-VAE KVs together, with source-VAE KV providing source-image details. These conditioning KVs are generated during prefill and reused throughout inference.

\paragraph{Candidate and protected segments.}
Beyond these conditioning KVs, the attention context contains objects with different management requirements. Boundary KVs often form attention sinks, while current-latent KV is recomputed at each denoising step, as illustrated in Figure~\ref{fig:fig_task_type_cache_distributions_L08}(a--b). Their distinct roles and update patterns motivate treating them separately as protected objects. 

\designprinciple{1}{
Identify the KV cache types present in each task and distinguish compression candidates from protected objects according to their roles and update patterns.
}

Accordingly, UniCache partitions the attention context into segments based on the current task and semantic role. The conditioning types identified above form the candidate segments passed to subsequent policy assignment or budget allocation, whereas boundary and current-latent KV remain uncompressed and are excluded from the shared compression budget.
\subsection{Offline Policy Assignment}
\begin{wrapfigure}{r}{0.45\textwidth}
	\centering
    \captionsetup{skip=1pt}
	\includegraphics[width=0.92\linewidth]{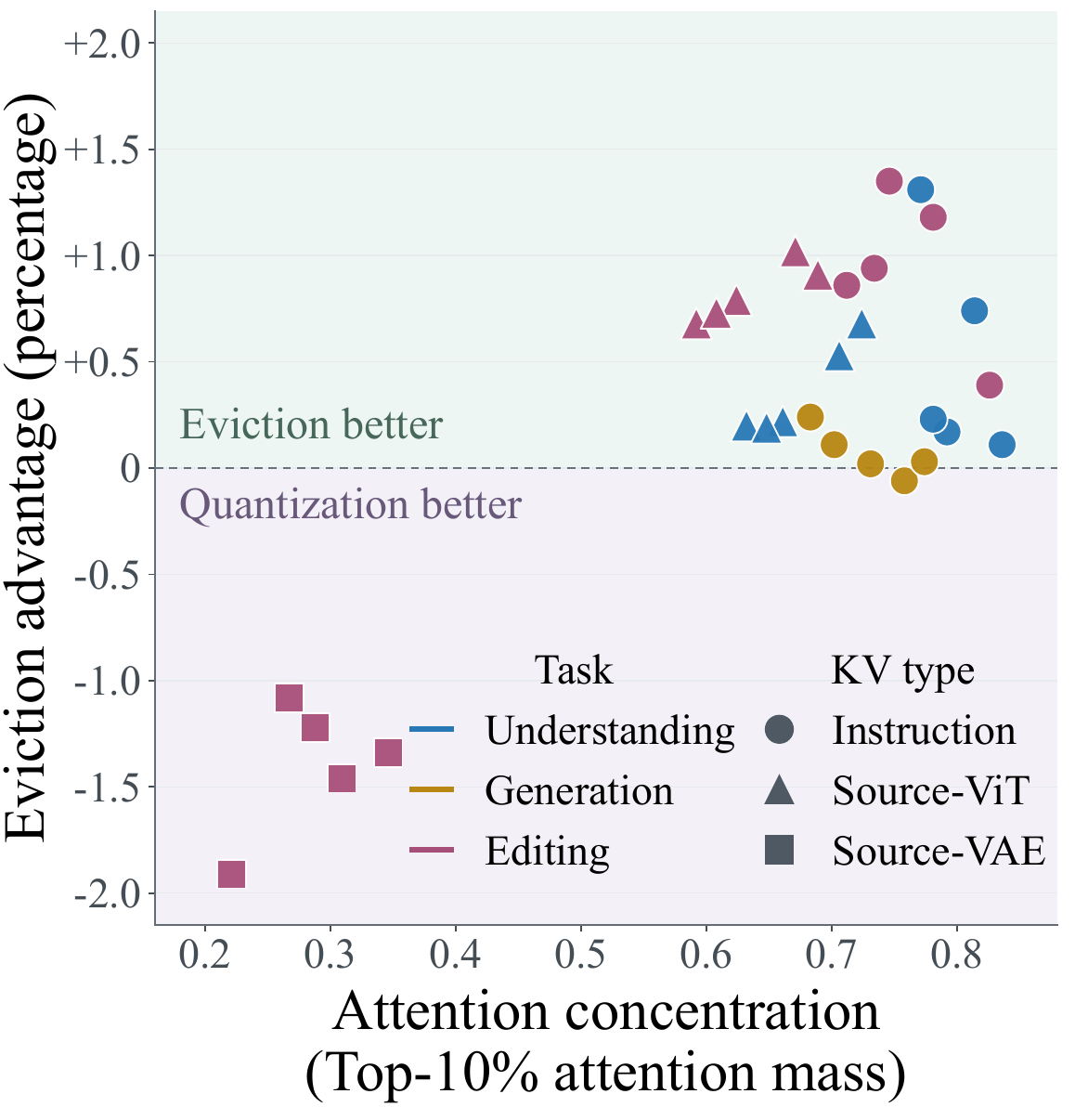}
	\caption{Each point represents a sampled subset for a task--KV-type pair with eviction or quantization at the same compression ratio. Details are provided in Appendix~\ref{app:policy_preference}.}
	\label{fig:sparsity_policy_preference}
\end{wrapfigure}
\paragraph{Type-dependent compression-policy preferences.}
The cache types present in a task can also favor different compression policies. As illustrated in Figure~\ref{fig:fig_task_type_cache_distributions_L08}(c--d), instruction and source-ViT KVs exhibit concentrated attention, which favors eviction policies that retain critical tokens while removing low-contribution tokens. Quantization, however, may perturb the semantic representations of critical instruction KVs. In contrast, source-VAE KVs exhibit broadly distributed attention, with lower attention probabilities for individual tokens, where eviction can discard source-image spatial details, whereas quantization preserves complete token coverage. These different sensitivities help explain the failure modes in Figure~\ref{fig:fig1_editing_comparison_extended}. Figure~\ref{fig:sparsity_policy_preference} also provides quantitative support for these policy preferences under matched compression budgets.

\designprinciple{2}{
 Compression policies should be assigned to cache types according to their attention distributions and compression sensitivity.
}
\par
\WFclear

\paragraph{Calibration-based policy assignment.}
UniCache assigns a compression policy to each candidate task--type pair using attention sparsity measured through offline calibration. These assignments remain fixed during inference, while their compression strengths are adjusted through online budget allocation.

\subsection{Attention-Guided Dynamic Budget Allocation}

\begin{wrapfigure}{r}{0.45\textwidth}
	\centering
    \captionsetup{skip=9pt}
	\includegraphics[width=0.99\linewidth]{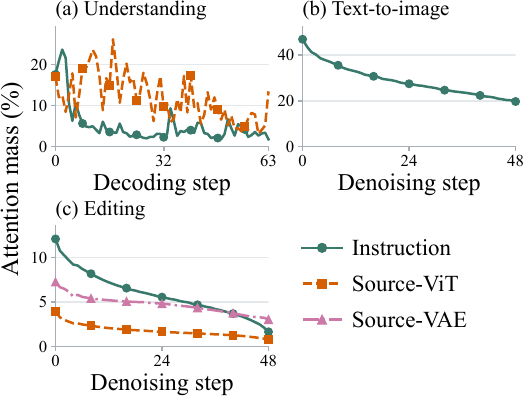}
	\caption{Attention mass across inference steps. Understanding exhibits step-to-step fluctuations, while generation and editing show overall declines at type-dependent rates.}
	\label{fig:fig_task_stage_attention_mass_L08_wrap}
\end{wrapfigure}

\looseness=-1 \paragraph{Unequal importance across cache types.} After assigning suitable compression policies to different cache types, we must also determine how much budget each policy should retain. As illustrated in Figure~\ref{fig:fig_task_type_cache_distributions_L08} and Figure~\ref{fig:fig_task_stage_attention_mass_L08_wrap}, different cache types receive different amounts of attention. A uniform retention ratio would allocate storage in proportion to the original cache sizes, without accounting for these differences. 

\designprinciple{3}{A global compression budget should be distributed across cache types according to their relative importance.}

\paragraph{Chunk-level attention statistics.}
UniCache estimates cache importance by aggregating attention probabilities over chunks of layers and inference steps, reducing scheduling frequency. For each candidate cache type, we sum the attention probabilities over its retained keys and average across heads, queries, and layer--step pairs:
\begin{equation}
m_a^{(c)}
=
\frac{1}{|\Omega_c|}
\sum_{(\ell,t)\in\Omega_c}
\operatorname{Mean}_{h,q}
\left[
\sum_{k\in\mathcal{R}_a^{\ell,t}}
P_{h,q,k}^{\ell,t}
\right].
\label{eq:chunk_attention_mass}
\end{equation}
Here, $\Omega_c$ contains the layer--step pairs in chunk $c$, and $\mathcal{R}_a^{\ell,t}$ contains the retained key indices of candidate cache type $a$ at layer $\ell$ and step $t$.
$P_{h,q,k}^{\ell,t}$ denotes the attention probability assigned by query $q$ in head $h$ to key $k$.
Statistics collected in chunk $c$ guide budget allocation for chunk $c+1$ within the same layer chunk.
We initialize $m_a^{(0)}$ from the final prefill queries for understanding and from the first full-KV denoising step for generation and editing.

\paragraph{Type-level budget allocation.}
Using these chunk-level attention statistics, UniCache distributes a shared storage budget across candidate cache types in proportion to their attention mass, capped by each type's previous allocation:
\begin{equation}
b_a^{(c+1)}
=
\min\left\{
b_a^{(c)},
\frac{m_a^{(c)}}{\sum_j m_j^{(c)}} B^{(c+1)}
\right\},
\qquad c\geq0.
\label{eq:type_budget}
\end{equation}
where $B^{(c+1)}$ denotes the shared storage budget, and $b_a^{(c)}$ denotes the allocated budget of type $a$ within the layer chunk.
We initialize $b_a^{(0)}=S_a$, where $S_a$ is the original storage cost of type $a$.
\looseness=-1 When a type reaches its previous allocation cap, remaining budget is redistributed proportionally among uncapped types with positive attention mass, with a small portion potentially remaining unused due to irreversible eviction.

\paragraph{Task- and stage-dependent cache importance.}
\looseness=-1 The allocation above determines how to divide a given budget, but the total budget should also vary across tasks or inference stages, as the model's reliance on conditioning KVs differs across these settings. As illustrated in Figure~\ref{fig:fig_task_stage_attention_mass_L08_wrap}, during understanding, attention contributions of conditional KVs  fluctuate with the newly generated tokens. In contrast, during generation and editing, the total attention mass assigned to conditioning KVs decreases as denoising progresses, and different cache types exhibit different rates of decline. Thus, the total budgets should also be temporally scheduled and adapt to task and inference stage.

\designprinciple{4}{The total budget of compression should be adjusted over time according to the task and inference stage.}
\par
\WFclear

\paragraph{Task-aware total budget.}
To implement this task-aware temporal scheduling of total budget, UniCache controls the total budget through a retention ratio $\rho^{(c+1)}$, giving $B^{(c+1)}=\rho^{(c+1)}\sum_a S_a$. Understanding uses a fixed task-level retention ratio because the conditioning context remains active throughout autoregressive decoding. For generation and editing, we instead adjust this ratio using the total candidate conditioning mass $M^{(c)}=\sum_a m_a^{(c)}$, relative to its initial full-KV reference $M^{(0)}>0$. The first compressed chunk uses $\rho^{(1)}=\rho_{\mathrm{init}}$, subsequent updates follow
\begin{equation}
\rho^{(c+1)}
=
\min\left\{
\rho^{(c)},
\max\left[
\rho_{\min},
\rho_{\mathrm{init}}
\frac{M^{(c)}}{M^{(0)}}
\right]
\right\},
\qquad c\geq1.
\label{eq:temporal_budget_schedule}
\end{equation}
Here, $0<\rho_{\min}\leq\rho_{\mathrm{init}}\leq1$ specify the minimum and initial retention ratios. Through this dynamic retention ratio, the global budget can be temporally scheduled with the relative importance of conditional KVs across tasks and inference stages.

\subsection{Type-Aware Parallel Compression}
\looseness=-1 Given these type-specific budgets, UniCache translates each allocation into the configuration of its assigned compression policy: token-eviction policies determine how many KV tokens to retain, while quantization policies determine the numerical precision and grouping granularity. Once configured, each policy maintains its own runtime state and compresses only its assigned segments, allowing the segment compression to execute independently and in parallel. After that, the compressed candidates and protected segments are then reassembled in their original token order for attention computation.

\section{Empirical Results}
\begin{table}[t]
	\centering
	\caption{\textbf{Cross-task results on BAGEL and SenseNova-U1.} MMBench reports circular accuracy (\%), MMMU reports validation accuracy (\%), and GEdit reports $Q_O$. Best and second-best compressed results within each model are \textbf{bold} and \underline{underlined}. }
	\label{tab:main_summary}
	\footnotesize
	\setlength{\tabcolsep}{2.5pt}
	\renewcommand{\arraystretch}{1.12}
	\resizebox{0.95\textwidth}{!}{%
		\begin{tabular}{llcccccccc}
			\toprule
			\multirow[c]{2}{*}[-1.7ex]{Model} & \multirow[c]{2}{*}[-1.7ex]{Method} & \multicolumn{3}{c}{Understanding} & Generation & \multicolumn{4}{c}{Editing} \\
			\cmidrule(lr){3-5}\cmidrule(lr){6-6}\cmidrule(lr){7-10}
			& & MME $\uparrow$ & MMBench $\uparrow$ & MMMU $\uparrow$ & GenEval $\uparrow$ & \shortstack{GEdit\\$Q_O\uparrow$} & \shortstack{PIE\\Structure $\downarrow$} & \shortstack{PIE\\PSNR $\uparrow$} & \shortstack{PIE CLIP\\Target $\uparrow$} \\
			\midrule
			\multirow{7}{*}{BAGEL} & - & 2373.21 & 85.14 & 53.33 & 0.781 & 6.947 & 0.101 & 18.805 & 28.174 \\
			& H$_2$O & \underline{2372.60} & \textbf{85.31} & \underline{52.78} & \underline{0.777} & 6.860 & 0.125 & 14.663 & 28.567 \\
			& KIVI & 2362.31 & 83.68 & 51.78 & \underline{0.777} & \underline{6.929} & \textbf{0.098} & \underline{18.576} & 27.812 \\
			& StreamingLLM & \underline{2372.60} & \textbf{85.31} & \underline{52.78} & 0.313 & 6.347 & 0.151 & 13.021 & \textbf{29.118} \\
			& SnapKV & 2371.01 & \underline{85.22} & 51.91 & 0.558 & 6.551 & 0.149 & 12.932 & \underline{28.942} \\
			& PyramidKV & 2367.02 & 84.88 & 52.12 & 0.558 & 6.325 & 0.153 & 12.549 & 28.912 \\
			& UniCache & \textbf{2377.36} & \textbf{85.31} & \textbf{52.89} & \textbf{0.780} & \textbf{6.965} & \underline{0.100} & \textbf{18.969} & 28.211 \\
			\midrule
			\multirow{4}{*}{\shortstack{SenseNova-U1}} & - & 2396.50 & 75.77 & 47.67 & 0.887 & 6.691 & 0.099 & 17.740 & 28.322 \\
			& H$_2$O & 2290.27 & \underline{75.43} & \underline{47.67} & 0.875 & \underline{4.977} & 0.142 & \underline{14.773} & 27.984 \\
			& KIVI & \underline{2377.65} & 74.66 & 47.00 & \underline{0.888} & 3.950 & \underline{0.131} & 14.750 & \underline{28.312} \\
			& UniCache & \textbf{2383.72} & \textbf{75.69} & \textbf{48.00} & \textbf{0.892} & \textbf{6.130} & \textbf{0.100} & \textbf{17.456} & \textbf{29.031} \\
			\bottomrule
		\end{tabular}
		
	}
\end{table}
\subsection{Experimental Setting}
\label{sec:experiment_setting}
\paragraph{Model and Benchmarks}
We evaluate UniCache on two representative MoT unified multimodal models BAGEL~\citep{deng2025emerging} and SenseNova-U1 ~\citep{diao2026sensenova}. We use MME~\citep{fu2026mme}, MMBench~\citep{liu2024mmbench}, and MMMU~\citep{yue2024mmmu} for understanding, GenEval~\citep{ghosh2023geneval} for text-to-image generation, and PIE-Bench~\citep{ju2023direct} and GEdit-Bench~\citep{liu2025step1x} with English instructions and Qwen2.5-VL-72B-Instruct~\citep{bai2025qwen25vltechnicalreport} as the judge for image editing. All experiments use the official evaluation protocols. 

\paragraph{Baselines}
We compare UniCache with the full-KV model and five representative KV cache compression methods: H$_2$O~\citep{zhang2023H2O}, KIVI~\citep{liu2024kivi}, StreamingLLM~\citep{xiao2024efficient}, SnapKV~\citep{li2024snapkv}, and PyramidKV~\citep{cai2024pyramidkv}. For controlled comparison, all methods are configured to achieve approximately matched KV-memory reductions: around $80\%$ on understanding and editing tasks and around $60\%$ on text-to-image generation. 

We also report the detailed settings of compression ratios and calibration in Appendix~\ref{app:detailed_setting} and our analysis of calibration statistics for SensoNova-U1 in Appendix~\ref{app:attn_analysis_sensenova}.

\subsection{Main Results}
As illustrated in Table~\ref{tab:main_summary}, UniCache effectively mitigates the failure modes encountered by prior KV cache compression methods and achieves balanced quality in all evaluated tasks through task-aware and type-aware KV cache compression.

For understanding tasks, methods that allocate a uniform compression budget introduce a bias between perceptual and cognitive capabilities. For instance, KIVI achieves the highest score in the Perception part in MME but the lowest in the Cognition part, as reported in Table~\ref{tab:bagel_mme_geneval}. By contrast, UniCache performs comparably to the original model on both dimensions.

For generation and editing, UniCache also retains nearly all of the original model's performance. For text-to-image generation, as illustrated in Table~\ref{tab:bagel_mme_geneval}, its adaptive budget allocation allows it to outperform existing methods in nearly every subcategory. For image editing, existing methods exhibit distinct limitations: token-eviction methods tend to discard structural details from the input image, whereas quantization methods compromise semantic understanding and instruction-following ability. Figure~\ref{fig:fig1_editing_comparison_extended} provides qualitative visual results. UniCache preserves different kinds of information and avoids these failure modes even under high compression ratios.

\begin{table}[t]
	\centering
	\caption{\textbf{Detailed MME and GenEval results on BAGEL.}
		Comp.\ denotes KV-cache compression.
		Higher is better for all quality metrics.
		Best and second-best results among compressed methods are
		\textbf{bold} and \underline{underlined}, respectively.}
	\label{tab:bagel_mme_geneval}
	\resizebox{\textwidth}{!}{%
		\renewcommand{\arraystretch}{1.15}
		\setlength{\tabcolsep}{3pt}
		\begin{tabular}{l*{14}{c}}
			\toprule
			\multirow{2}{*}{Method}
			& \multicolumn{4}{c}{MME}
			& \multicolumn{10}{c}{GenEval} \\
			\cmidrule(lr){2-5}\cmidrule(lr){6-15}
			& Comp. & Perception & Cognition & Total
			& Comp. & Single & Two & Count & Color
			& Position & Binding & \shortstack{Image\\Acc.}
			& \shortstack{Prompt\\Acc.} & Overall \\
			\midrule
			BF16
			& 0\% & 1680.35 & 692.86 & 2373.21
			& 0\% & 0.997 & 0.922 & 0.781 & 0.846
			& 0.520 & 0.623 & 0.773 & 0.897 & 0.781 \\
			H$_2$O
			& 80.07\% & 1679.74 & \underline{692.86}
			& \underline{2372.60}
			& 59.47\% & \underline{0.997} & 0.909
			& \textbf{0.788} & \textbf{0.864}
			& \underline{0.505} & 0.600
			& \underline{0.768} & \underline{0.902}
			& \underline{0.777} \\
			KIVI
			& 80.69\% & \textbf{1685.53} & 676.79 & 2362.31
			& 59.12\% & \textbf{1.000} & \underline{0.912}
			& \underline{0.778} & 0.840
			& \underline{0.505} & \textbf{0.628}
			& \underline{0.768} & 0.897 & \underline{0.777} \\
			StreamingLLM
			& 79.94\% & 1679.74 & \underline{692.86}
			& \underline{2372.60}
			& 58.01\% & 0.863 & 0.051 & 0.419 & 0.540
			& 0.003 & 0.003 & 0.287 & 0.405 & 0.313 \\
			SnapKV
			& 79.94\% & 1679.58 & 691.43 & 2371.01
			& 61.25\% & \underline{0.997} & 0.503 & 0.659 & 0.832
			& 0.095 & 0.263 & 0.536 & 0.673 & 0.558 \\
			PyramidKV
			& 79.94\% & 1676.30 & 690.71 & 2367.02
			& 61.25\% & \underline{0.997} & 0.505 & 0.659 & 0.832
			& 0.095 & 0.260 & 0.536 & 0.675 & 0.558 \\
			\midrule
			UniCache
			& 80.00\% & \underline{1681.65}
			& \textbf{695.71} & \textbf{2377.36}
			& 59.76\% & \textbf{1.000} & \textbf{0.929}
			& 0.763 & \textbf{0.864}
			& \textbf{0.515} & \underline{0.608}
			& \textbf{0.771} & \textbf{0.906} & \textbf{0.780} \\
			\bottomrule
		\end{tabular}%
	}
\end{table}

\subsection{Ablation Studies}
\label{sec:ablation_studies}
\paragraph{Component Analysis}
\looseness=-1 To assess the contribution of each component, we conduct ablation studies under a consistent experimental setup. Specifically, we remove each component from UniCache individually and evaluate the resulting variants across several benchmarks. As shown in Table \ref{tab:component_summary},  removing the protection of boundary and current-latent KV leads to degraded text-to-image generation performance. Moreover, attention-guided allocation is particularly critical for image understanding and image editing which involve multiple KV-cache types. Given the same total cache budget, the task-aware temporal schedule allocates more KV cache capacity to the early stages of generation and less to the later stages, so removing it causes slightly degradation in generation tasks as the uniform budget allocation across steps does not track the decline in attention mass of conditional KV across denoising steps. Collectively, these components enable UniCache to deliver balanced performance across tasks. 
\begin{table}[t]
    \centering
    \caption{\textbf{Method component ablations across tasks.} Targets are approximately $80\%$ compression on MME and $60\%$ on GenEval and PIE-Bench. Dashes indicate inapplicable settings.}
    \label{tab:component_summary}
    \small
    \setlength{\tabcolsep}{3pt}
    \resizebox{0.9\textwidth}{!}{%
    \begin{tabular}{p{130pt}ccccccc}
        \toprule
        & \multicolumn{3}{c}{MME}
        & GenEval
        & \multicolumn{3}{c}{PIE-Bench} \\
        \cmidrule(lr){2-4}\cmidrule(lr){5-5}\cmidrule(lr){6-8}
        Setting
        & Perception $\uparrow$
        & Cognition $\uparrow$
        & Total $\uparrow$
        & Overall $\uparrow$
        & Structure $\downarrow$
        & PSNR $\uparrow$
        & \shortstack{CLIP\\Target $\uparrow$} \\
        \midrule
        BAGEL
        & 1680.35 & 692.86 & 2373.21
        & 0.781 & 0.101 & 18.805 & 28.174 \\
        Full UniCache
        & 1681.65 & 695.71 & 2377.36
        & 0.780 & 0.101 & 19.100 & 28.171 \\
        w/o Task/Type Awareness
        & 1679.74 & 692.86 & 2372.60
        & 0.767 & 0.103 & 17.875 & 28.221 \\
        w/o Protection
        & 1680.47 & 692.86 & 2373.32
        & 0.567 & 0.102 & 18.959 & 27.235 \\
        w/o Attention-Guided Allocation
        & 1680.32 & 688.01 & 2368.33
        & --- & 0.121 & 17.277 & 27.036 \\
        w/o Task-Aware Temporal Schedule
        & --- & --- & ---
        & 0.766 & 0.110 & 18.477 & 28.010 \\
        \bottomrule
    \end{tabular}%
    }
\end{table}
\paragraph{Temporal schedule.}
We further compare attention-guided scheduling against two alternatives: a constant budget and a linear decay schedule. The former maintains a fixed retention ratio throughout denoising, whereas the latter linearly interpolates from $\rho_{\mathrm{init}}$ to $\rho_{\min}$ as denoising progresses. All other components and compression policies are held fixed as described in Section \ref{sec:experiment_setting}. As shown in Table \ref{tab:temporal_schedule}, attention-guided decay matches linear decay on GenEval and achieves the best results across all PIE-Bench metrics, showing that conditioning attention provides a more effective temporal signal for budget allocation across steps.
\begin{table}
	\centering
	\caption{\textbf{Ablation of temporal budget schedules.} We change the schedule methods for global budget scheduling and report results in GenEval and PIE-Bench.}
	\label{tab:temporal_schedule}
	\resizebox{0.9\columnwidth}{!}{
		\begin{tabular}{lcccccc}
			\toprule
			\textbf{Schedule}
			& \textbf{GenEval $\uparrow$}
			& \textbf{Structure $\downarrow$}
			& \textbf{LPIPS $\downarrow$}
			& \textbf{SSIM $\uparrow$}
			& \textbf{CLIP Target $\uparrow$}
			& \textbf{CLIP Edit $\uparrow$} \\
			\midrule
			Constant
			& 0.776 & 0.102 & 0.184 & 0.757 & 28.010 & 27.070 \\
			Linear decay
			& \textbf{0.780} & 0.102 & 0.184 & 0.757
			& 28.187 & 27.285 \\
			Attention-guided decay
			& \textbf{0.780} & \textbf{0.101} & \textbf{0.179}
			& \textbf{0.763} & \textbf{28.211} & \textbf{27.293} \\
			\bottomrule
		\end{tabular}
	}
\end{table}
\begin{figure}[t]
    \centering
    \includegraphics[width=0.99\linewidth]{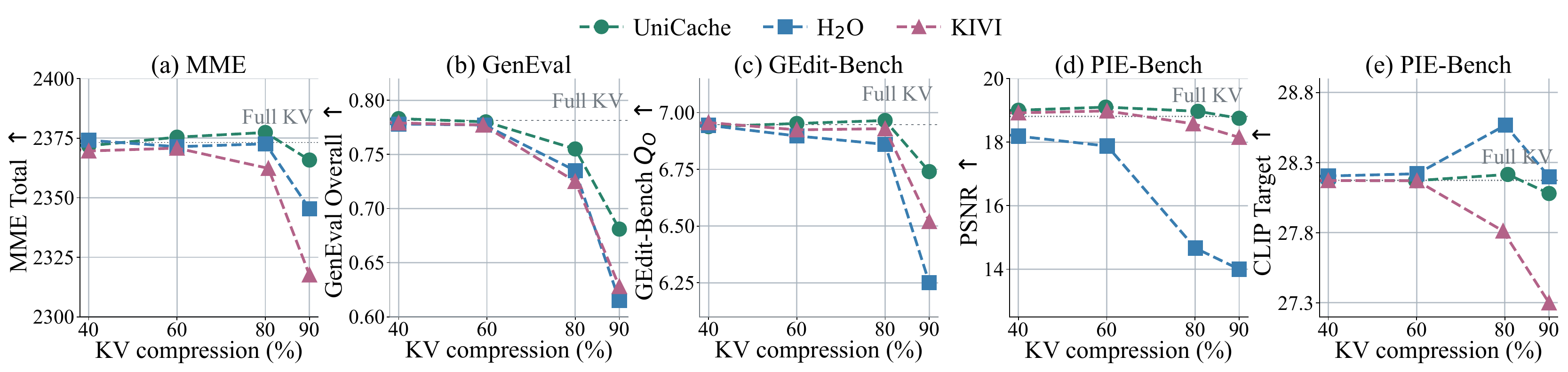}
    \caption{\textbf{Ablation of compression ratio across different methods and benchmarks.} Across different benchmarks, UniCache is robust to high compression ratio and maintains balanced performance at very high compression ratios.}
    \label{fig:five_panel_comparison}
\end{figure}
\begin{figure}[H]
	\centering
	\includegraphics[width=0.95\textwidth]{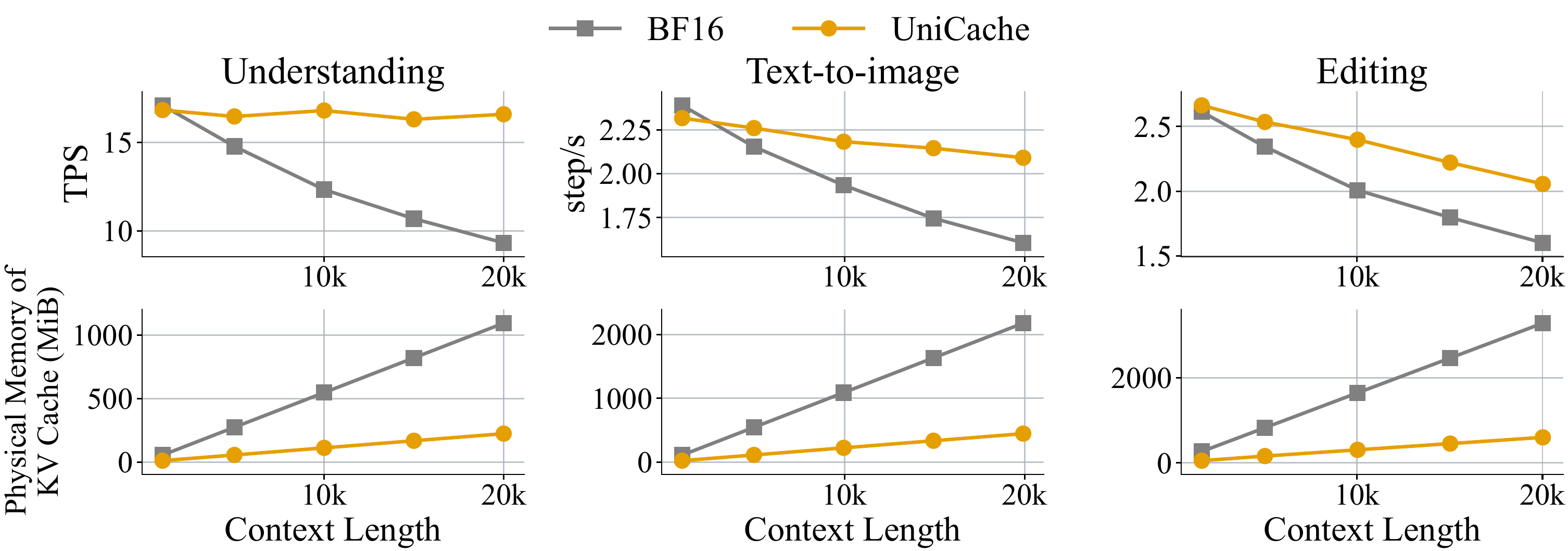}
	\caption{\textbf{Runtime efficiency of UniCache across context lengths.}
		Throughput is measured in tokens per second for understanding and denoising steps per second for generation and editing. Memory reports the persistent managed KV cache memory summed over all layers.}
	\label{fig:efficiency_throughput}
\end{figure}
\paragraph{Compression ratio.} We also compare UniCache with KIVI and H${}_2$O in different compression ratios and report the results in Figure~\ref{fig:five_panel_comparison}. Across different benchmarks, UniCache is more robust to high compression ratio than other methods. Notably, in PIE-Bench, H${}_2$O and KIVI cannot avoid degradation in at least one of the two metrics, while UniCache maintains balanced performance across both metrics.
\subsection{End-to-End Efficiency}
\looseness=-1 We implement an efficient framework with physical KV compaction and low-bit memory, as introduced in Appendix \ref{app:open_source}.  All measurements are conducted on a single NVIDIA A100 GPU (40 GB). We report detailed results of BAGEL-7B-MoT in Figure~\ref{fig:efficiency_throughput}. The results show that the benefit of physical compression increases with context length. At approximately 20K KV tokens per attention branch, UniCache improves throughput by up to $1.78\times$ while reducing persistent managed KV cache memory by about $80\%$.

\section{Conclusion}
\label{sec:conclusion}
	\looseness=-1 In this work, we show that heterogeneous cache types and task-dependent dynamics limit existing KV-cache compression methods in MoT-based unified multimodal models. To address this challenge, we introduce UniCache, a training-free framework that combines offline type-aware policy assignment with attention-guided budget allocation and task-aware temporal scheduling. By coordinating different compression policies across cache segments, UniCache preserves quality across understanding, generation, and editing, achieving up to $5\times$ KV-cache compression with negligible performance loss and up to $1.78\times$ long-context throughput, improving the practicality of scaling unified multimodal model to longer context.

\label{page:references}
\bibliographystyle{iclr2027_conference}
\bibliography{iclr2027_conference}

\begin{thebibliography}{28}
\providecommand{\natexlab}[1]{#1}
\providecommand{\url}[1]{\texttt{#1}}
\expandafter\ifx\csname urlstyle\endcsname\relax
  \providecommand{\doi}[1]{doi: #1}\else
  \providecommand{\doi}{doi: \begingroup \urlstyle{rm}\Url}\fi

\bibitem[Bai et~al.(2025)Bai, Chen, Liu, Wang, Ge, Song, Dang, Wang, Wang,
  Tang, Zhong, Zhu, Yang, Li, Wan, Wang, Ding, Fu, Xu, Ye, Zhang, Xie, Cheng,
  Zhang, Yang, Xu, and Lin]{bai2025qwen25vltechnicalreport}
Shuai Bai, Keqin Chen, Xuejing Liu, Jialin Wang, Wenbin Ge, Sibo Song, Kai
  Dang, Peng Wang, Shijie Wang, Jun Tang, Humen Zhong, Yuanzhi Zhu, Mingkun
  Yang, Zhaohai Li, Jianqiang Wan, Pengfei Wang, Wei Ding, Zheren Fu, Yiheng
  Xu, Jiabo Ye, Xi~Zhang, Tianbao Xie, Zesen Cheng, Hang Zhang, Zhibo Yang,
  Haiyang Xu, and Junyang Lin.
\newblock Qwen2.5-vl technical report, 2025.
\newblock URL \url{https://arxiv.org/abs/2502.13923}.

\bibitem[Cai et~al.(2025)Cai, Zhang, Gao, Liu, Li, Liu, Lu, Xiong, Dong, Hu,
  and Xiao]{cai2024pyramidkv}
Zefan Cai, Yichi Zhang, Bofei Gao, Yuliang Liu, Yucheng Li, Tianyu Liu, Keming
  Lu, Wayne Xiong, Yue Dong, Junjie Hu, and Wen Xiao.
\newblock Pyramidkv: Dynamic kv cache compression based on pyramidal
  information funneling, 2025.
\newblock URL \url{https://arxiv.org/abs/2406.02069}.

\bibitem[Deng et~al.(2025)Deng, Zhu, Li, Gou, Li, Wang, Zhong, Yu, Nie, Song,
  et~al.]{deng2025emerging}
Chaorui Deng, Deyao Zhu, Kunchang Li, Chenhui Gou, Feng Li, Zeyu Wang, Shu
  Zhong, Weihao Yu, Xiaonan Nie, Ziang Song, et~al.
\newblock Emerging properties in unified multimodal pretraining.
\newblock \emph{arXiv preprint arXiv:2505.14683}, 2025.

\bibitem[Diao et~al.(2026)Diao, Wu, Deng, Wang, Bai, Wu, Fan, Ye, Tong, Fan,
  Li, Wang, Cao, Lin, Yang, Cai, Niu, Zhu, Liu, Lv, Yu, Xie, Wang, Fan, Li, Lu,
  Ni, Xu, Liang, Shi, Dai, Wang, Qian, Gao, Liu, Sun, Shen, Wang, Ma, Yang,
  Xie, Li, Zhong, Kong, Shi, Gao, Yao, Wang, Bai, Lin, Yin, Sun, Gong, Wang,
  Lu, Yang, Liu, and Lin]{diao2026sensenova}
Haiwen Diao, Penghao Wu, Hanming Deng, Jiahao Wang, Shihao Bai, Silei Wu,
  Weichen Fan, Wenjie Ye, Wenwen Tong, Xiangyu Fan, Yan Li, Yubo Wang, Zhijie
  Cao, Zhiqian Lin, Zhitao Yang, Zhongang Cai, Yuwei Niu, Yue Zhu, Bo~Liu,
  Chengguang Lv, Haojia Yu, Haozhe Xie, Hongli Wang, Jianan Fan, Jiaqi Li,
  Jiefan Lu, Jingcheng Ni, Junxiang Xu, Kaihuan Liang, Lianqiang Shi, Linjun
  Dai, Linyan Wang, Oscar Qian, Peng Gao, Pengfei Liu, Qingping Sun, Rui Shen,
  Ruisi Wang, Shengnan Ma, Shuang Yang, Siyi Xie, Siying Li, Tianbo Zhong,
  Xiangli Kong, Xuanke Shi, Yang Gao, Yongqiang Yao, Yves Wang, Zhengqi Bai,
  Zhengyu Lin, Zixin Yin, Wenxiu Sun, Ruihao Gong, Quan Wang, Lewei Lu, Lei
  Yang, Ziwei Liu, and Dahua Lin.
\newblock Sensenova-u1: Unifying multimodal understanding and generation with
  neo-unify architecture, 2026.
\newblock URL \url{https://arxiv.org/abs/2605.12500}.

\bibitem[Fu et~al.(2025)Fu, Chen, Shen, Qin, Zhang, Lin, Yang, Zheng, Li, Sun,
  Wu, Ji, Shan, and He]{fu2026mme}
Chaoyou Fu, Peixian Chen, Yunhang Shen, Yulei Qin, Mengdan Zhang, Xu~Lin,
  Jinrui Yang, Xiawu Zheng, Ke~Li, Xing Sun, Yunsheng Wu, Rongrong Ji, Caifeng
  Shan, and Ran He.
\newblock Mme: A comprehensive evaluation benchmark for multimodal large
  language models, 2025.
\newblock URL \url{https://arxiv.org/abs/2306.13394}.

\bibitem[Ghosh et~al.(2023)Ghosh, Hajishirzi, and Schmidt]{ghosh2023geneval}
Dhruba Ghosh, Hannaneh Hajishirzi, and Ludwig Schmidt.
\newblock Geneval: An object-focused framework for evaluating text-to-image
  alignment.
\newblock \emph{Advances in Neural Information Processing Systems},
  36:\penalty0 52132--52152, 2023.

\bibitem[He et~al.(2024)He, Chen, Liu, Shao, Zhou, Zhang, and
  Zhuang]{he2024zipvl}
Yefei He, Feng Chen, Jing Liu, Wenqi Shao, Hong Zhou, Kaipeng Zhang, and Bohan
  Zhuang.
\newblock Zipvl: Efficient large vision-language models with dynamic token
  sparsification.
\newblock \emph{arXiv preprint arXiv:2410.08584}, 2024.

\bibitem[Ju et~al.(2023)Ju, Zeng, Bian, Liu, and Xu]{ju2023direct}
Xuan Ju, Ailing Zeng, Yuxuan Bian, Shaoteng Liu, and Qiang Xu.
\newblock Direct inversion: Boosting diffusion-based editing with 3 lines of
  code.
\newblock \emph{arXiv preprint arXiv:2310.01506}, 2023.

\bibitem[Ke et~al.(2026)Ke, Wen, Yang, Yang, Liu, Liao, Chen, Wang, and
  Zhang]{ke2026flash}
Junlong Ke, Zichen Wen, Boxue Yang, Yantai Yang, Xuyang Liu, Chenfei Liao,
  Zhaorun Chen, Shaobo Wang, and Linfeng Zhang.
\newblock Flash-unified: A training-free and task-aware acceleration framework
  for native unified models, 2026.
\newblock URL \url{https://arxiv.org/abs/2603.15271}.

\bibitem[Kwon et~al.(2023)Kwon, Li, Zhuang, Sheng, Zheng, Yu, Gonzalez, Zhang,
  and Stoica]{kwon2023efficient}
Woosuk Kwon, Zhuohan Li, Siyuan Zhuang, Ying Sheng, Lianmin Zheng, Cody~Hao Yu,
  Joseph Gonzalez, Hao Zhang, and Ion Stoica.
\newblock Efficient memory management for large language model serving with
  pagedattention.
\newblock In \emph{Proceedings of the 29th symposium on operating systems
  principles}, pp.\  611--626, 2023.

\bibitem[Li et~al.(2024)Li, Huang, Yang, Venkitesh, Locatelli, Ye, Cai, Lewis,
  and Chen]{li2024snapkv}
Yuhong Li, Yingbing Huang, Bowen Yang, Bharat Venkitesh, Acyr Locatelli,
  Hanchen Ye, Tianle Cai, Patrick Lewis, and Deming Chen.
\newblock Snapkv: Llm knows what you are looking for before generation, 2024.
\newblock URL \url{https://arxiv.org/abs/2404.14469}.

\bibitem[Liang et~al.(2024)Liang, Yu, Luo, Iyer, Dong, Zhou, Ghosh, Lewis, Yih,
  Zettlemoyer, et~al.]{liang2024mixture}
Weixin Liang, Lili Yu, Liang Luo, Srinivasan Iyer, Ning Dong, Chunting Zhou,
  Gargi Ghosh, Mike Lewis, Wen-tau Yih, Luke Zettlemoyer, et~al.
\newblock Mixture-of-transformers: A sparse and scalable architecture for
  multi-modal foundation models.
\newblock \emph{arXiv preprint arXiv:2411.04996}, 2024.

\bibitem[Liu et~al.(2025)Liu, Han, Xing, Yin, Wang, Cheng, Liao, Wang, Fu, Han,
  et~al.]{liu2025step1x}
Shiyu Liu, Yucheng Han, Peng Xing, Fukun Yin, Rui Wang, Wei Cheng, Jiaqi Liao,
  Yingming Wang, Honghao Fu, Chunrui Han, et~al.
\newblock Step1x-edit: A practical framework for general image editing.
\newblock \emph{arXiv preprint arXiv:2504.17761}, 2025.

\bibitem[Liu et~al.(2024{\natexlab{a}})Liu, Duan, Zhang, Li, Zhang, Zhao, Yuan,
  Wang, He, Liu, et~al.]{liu2024mmbench}
Yuan Liu, Haodong Duan, Yuanhan Zhang, Bo~Li, Songyang Zhang, Wangbo Zhao, Yike
  Yuan, Jiaqi Wang, Conghui He, Ziwei Liu, et~al.
\newblock Mmbench: Is your multi-modal model an all-around player?
\newblock In \emph{European conference on computer vision}, pp.\  216--233.
  Springer, 2024{\natexlab{a}}.

\bibitem[Liu et~al.(2023)Liu, Desai, Liao, Wang, Xie, Xu, Kyrillidis, and
  Shrivastava]{liu2023scissorhands}
Zichang Liu, Aditya Desai, Fangshuo Liao, Weitao Wang, Victor Xie, Zhaozhuo Xu,
  Anastasios Kyrillidis, and Anshumali Shrivastava.
\newblock Scissorhands: Exploiting the persistence of importance hypothesis for
  llm kv cache compression at test time, 2023.
\newblock URL \url{https://arxiv.org/abs/2305.17118}.

\bibitem[Liu et~al.(2024{\natexlab{b}})Liu, Yuan, Jin, Zhong, Xu, Braverman,
  Chen, and Hu]{liu2024kivi}
Zirui Liu, Jiayi Yuan, Hongye Jin, Shaochen Zhong, Zhaozhuo Xu, Vladimir
  Braverman, Beidi Chen, and Xia Hu.
\newblock Kivi: A tuning-free asymmetric 2bit quantization for kv cache.
\newblock \emph{arXiv preprint arXiv:2402.02750}, 2024{\natexlab{b}}.

\bibitem[Lu et~al.(2025)Lu, Xia, Zhang, Kuang, Zheng, Ren, and
  Xiao]{lu2025hyperbagel}
Yanzuo Lu, Xin Xia, Manlin Zhang, Huafeng Kuang, Jianbin Zheng, Yuxi Ren, and
  Xuefeng Xiao.
\newblock Hyper-bagel: A unified acceleration framework for multimodal
  understanding and generation, 2025.
\newblock URL \url{https://arxiv.org/abs/2509.18824}.

\bibitem[Shazeer(2019)]{shazeer2019fast}
Noam Shazeer.
\newblock Fast transformer decoding: One write-head is all you need.
\newblock \emph{arXiv preprint arXiv:1911.02150}, 2019.

\bibitem[Team(2024)]{team2024chameleon}
Chameleon Team.
\newblock Chameleon: Mixed-modal early-fusion foundation models.
\newblock \emph{arXiv preprint arXiv:2405.09818}, 2024.

\bibitem[Tu et~al.(2025)Tu, Vashchilenko, Lu, and Xu]{tu2025vl}
Dezhan Tu, Danylo Vashchilenko, Yuzhe Lu, and Panpan Xu.
\newblock Vl-cache: Sparsity and modality-aware kv cache compression for
  vision-language model inference acceleration.
\newblock In \emph{International Conference on Learning Representations},
  volume 2025, pp.\  219--239, 2025.

\bibitem[Wu et~al.(2025)Wu, Chen, Wu, Ma, Liu, Pan, Liu, Xie, Yu, Ruan,
  et~al.]{wu2025janus}
Chengyue Wu, Xiaokang Chen, Zhiyu Wu, Yiyang Ma, Xingchao Liu, Zizheng Pan, Wen
  Liu, Zhenda Xie, Xingkai Yu, Chong Ruan, et~al.
\newblock Janus: Decoupling visual encoding for unified multimodal
  understanding and generation.
\newblock In \emph{2025 IEEE/CVF Conference on Computer Vision and Pattern
  Recognition (CVPR)}, pp.\  12966--12977. IEEE, 2025.

\bibitem[Xiao et~al.(2024)Xiao, Tian, Chen, Han, and Lewis]{xiao2024efficient}
Guangxuan Xiao, Yuandong Tian, Beidi Chen, Song Han, and Mike Lewis.
\newblock Efficient streaming language models with attention sinks, 2024.
\newblock URL \url{https://arxiv.org/abs/2309.17453}.

\bibitem[Xie et~al.(2025)Xie, Mao, Bai, Zhang, Wang, Lin, Gu, Chen, Yang, and
  Shou]{xie2025show}
Jinheng Xie, Weijia Mao, Zechen Bai, David~Junhao Zhang, Weihao Wang,
  Kevin~Qinghong Lin, Yuchao Gu, Zhijie Chen, Zhenheng Yang, and Mike~Zheng
  Shou.
\newblock Show-o: One single transformer to unify multimodal understanding and
  generation.
\newblock In \emph{International Conference on Learning Representations},
  volume 2025, pp.\  28240--28264, 2025.

\bibitem[Yang et~al.(2026)Yang, Ma, Conzelmann, Zheng, Mahoney, Rusch, and
  Liu]{yang2026alphaq}
Wanqi Yang, Yuexiao Ma, Alexander Conzelmann, Xiawu Zheng, Michael~W Mahoney,
  T~Konstantin Rusch, and Shiwei Liu.
\newblock Alphaq: Calibration-free bit allocation for mixture-of-experts
  quantization.
\newblock \emph{arXiv preprint arXiv:2606.04980}, 2026.

\bibitem[Yue et~al.(2024)Yue, Ni, Zhang, Zheng, Liu, Zhang, Stevens, Jiang,
  Ren, Sun, et~al.]{yue2024mmmu}
Xiang Yue, Yuansheng Ni, Kai Zhang, Tianyu Zheng, Ruoqi Liu, Ge~Zhang, Samuel
  Stevens, Dongfu Jiang, Weiming Ren, Yuxuan Sun, et~al.
\newblock Mmmu: A massive multi-discipline multimodal understanding and
  reasoning benchmark for expert agi.
\newblock In \emph{Proceedings of the IEEE/CVF conference on computer vision
  and pattern recognition}, pp.\  9556--9567, 2024.

\bibitem[Zhang et~al.(2025)Zhang, Hu, Zhao, Lui, and Chen]{zhang2025diffkv}
Yanqi Zhang, Yuwei Hu, Runyuan Zhao, John~CS Lui, and Haibo Chen.
\newblock Diffkv: Differentiated memory management for large language models
  with parallel kv compaction.
\newblock In \emph{Proceedings of the ACM SIGOPS 31st Symposium on Operating
  Systems Principles}, pp.\  431--445, 2025.

\bibitem[Zhang et~al.(2023)Zhang, Sheng, Zhou, Chen, Zheng, Cai, Song, Tian,
  R{\'e}, Barrett, et~al.]{zhang2023H2O}
Zhenyu Zhang, Ying Sheng, Tianyi Zhou, Tianlong Chen, Lianmin Zheng, Ruisi Cai,
  Zhao Song, Yuandong Tian, Christopher R{\'e}, Clark Barrett, et~al.
\newblock H2o: Heavy-hitter oracle for efficient generative inference of large
  language models.
\newblock \emph{Advances in neural information processing systems},
  36:\penalty0 34661--34710, 2023.

\bibitem[Zhou et~al.(2025)Zhou, Yu, Babu, Tirumala, Yasunaga, Shamis, Kahn, Ma,
  Zettlemoyer, and Levy]{zhou2025transfusion}
Chunting Zhou, Lili Yu, Arun Babu, Kushal Tirumala, Michihiro Yasunaga, Leonid
  Shamis, Jacob Kahn, Xuezhe Ma, Luke Zettlemoyer, and Omer Levy.
\newblock Transfusion: Predict the next token and diffuse images with one
  multi-modal model.
\newblock In \emph{International Conference on Learning Representations},
  volume 2025, pp.\  6446--6469, 2025.

\end{thebibliography}

\clearpage
\appendix
\section{Attention Concentration and Policy Preference Experiments}
\label{app:policy_preference}

To measuring the relationship between attention concentration and policy preference, for each task--KV-type pair, we evaluate five sampled subsets, each containing 100 examples from the corresponding benchmark MME, GenEval and PIE-Bench. We compare H$_2$O  and KIVI as representative methods of eviction and quantization under matched KV-memory budgets, applying compression only to the selected KV type and leaving the remaining types unchanged. Each compressed run is evaluated against its corresponding  full-KV baseline using the task's primary metric.

To compare different policies, the metric we use to evaluate relative quality degradation is
$D_{\pi}=100(s_{\mathrm{full}}-s_{\pi})/s_{\mathrm{full}}$. The numerator is reversed for a lower-is-better metric.
Then we define eviction advantage as $D_{\mathrm{KIVI}}-D_{\mathrm{H_2O}}$, measured in percentage
points. As a result, positive values favor eviction and negative values favor quantization.  We report results in Figure~\ref{fig:sparsity_policy_preference}.

\section{Detailed Experimental Settings}
\label{app:detailed_setting}
\paragraph{Compression Ratio}
For each experiment, we set a target compression ratio and record the realized compression ratio, that the method actually achieves. We report the target and realized KV cache compression ratio in Table \ref{tab:compression_rates}.
\begin{table}[ht]
	\centering
	\caption{Target and realized KV cache compression
		ratios (\%) on BAGEL. }
	\label{tab:compression_rates}
	\small
	\setlength{\tabcolsep}{12pt}
	\begin{tabular}{lccc}
		\toprule
		\textbf{Method}
		& \textbf{Understanding}
		& \textbf{Text-to-Image}
		& \textbf{Image Editing} \\
		\midrule
		\textbf{Target compression}
		& \textbf{80.00}
		& \textbf{60.00}
		& \textbf{80.00} \\
		\midrule
		H$_2$O       & 80.07 & 59.47 & 80.09 \\
		KIVI         & 80.69 & 59.12 & 79.59 \\
		StreamingLLM & 79.94 & 58.01 & 79.94 \\
		SnapKV       & 79.94 & 61.25 & 79.93 \\
		PyramidKV    & 79.94 & 61.25 & 79.93 \\
		UniCache     & 80.00 & 59.76 & 80.67 \\
		\bottomrule
	\end{tabular}
\end{table}
\paragraph{Calibration Setting}
We use calibration to determine compression policy. For each task, we randomly sample 10 calibration examples and measure attention concentration as the fraction of within-type attention mass captured by the top 10\% of tokens, averaged across examples. We assign eviction to task–type pairs with concentration at least 0.5 and quantization otherwise, while calibrating \(\rho_{\min}\) to match the target average KV compression ratio. 

\section{Attention Analysis of SenseNova-U1}
\label{app:attn_analysis_sensenova}
\begin{figure}[t]
	\centering
	\includegraphics[width=\textwidth]{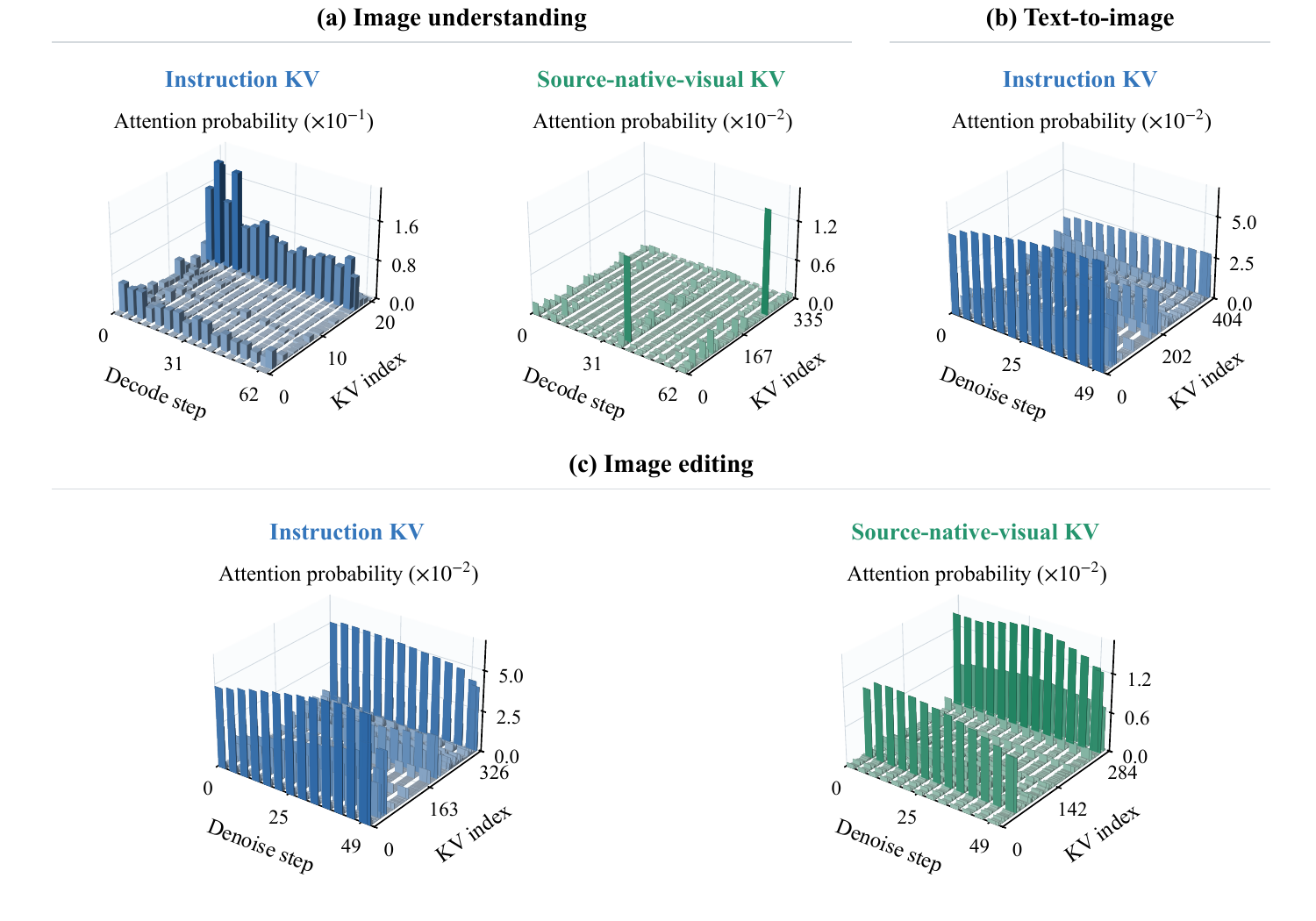}
	\caption{\textbf{Task- and type-dependent KV-cache attention patterns in SenseNova-U1 at layer 8.} Each panel shows token-level attention probability averaged over heads and queries for one example, with tokens kept in their original order. Source-native-visual KV denotes SenseNova-U1's native source-image representation.}
	\label{fig:sensenova_analysis}
\end{figure}

Unlike BAGEL, SenseNova-U1 represents the source image with a single native visual KV type. As shown in Figure~\ref{fig:sensenova_analysis}, our offline calibration finds that attention over this KV type is also concentrated on a small subset of tokens. UniCache therefore automatically assigns an eviction policy to source-native-visual KV, adapting its policy assignment to the model's cache structure and attention distribution. 

\section{Sampled Results Generated by UniCache}
\begin{figure}[t]
    \centering
    \includegraphics[width=1\linewidth]{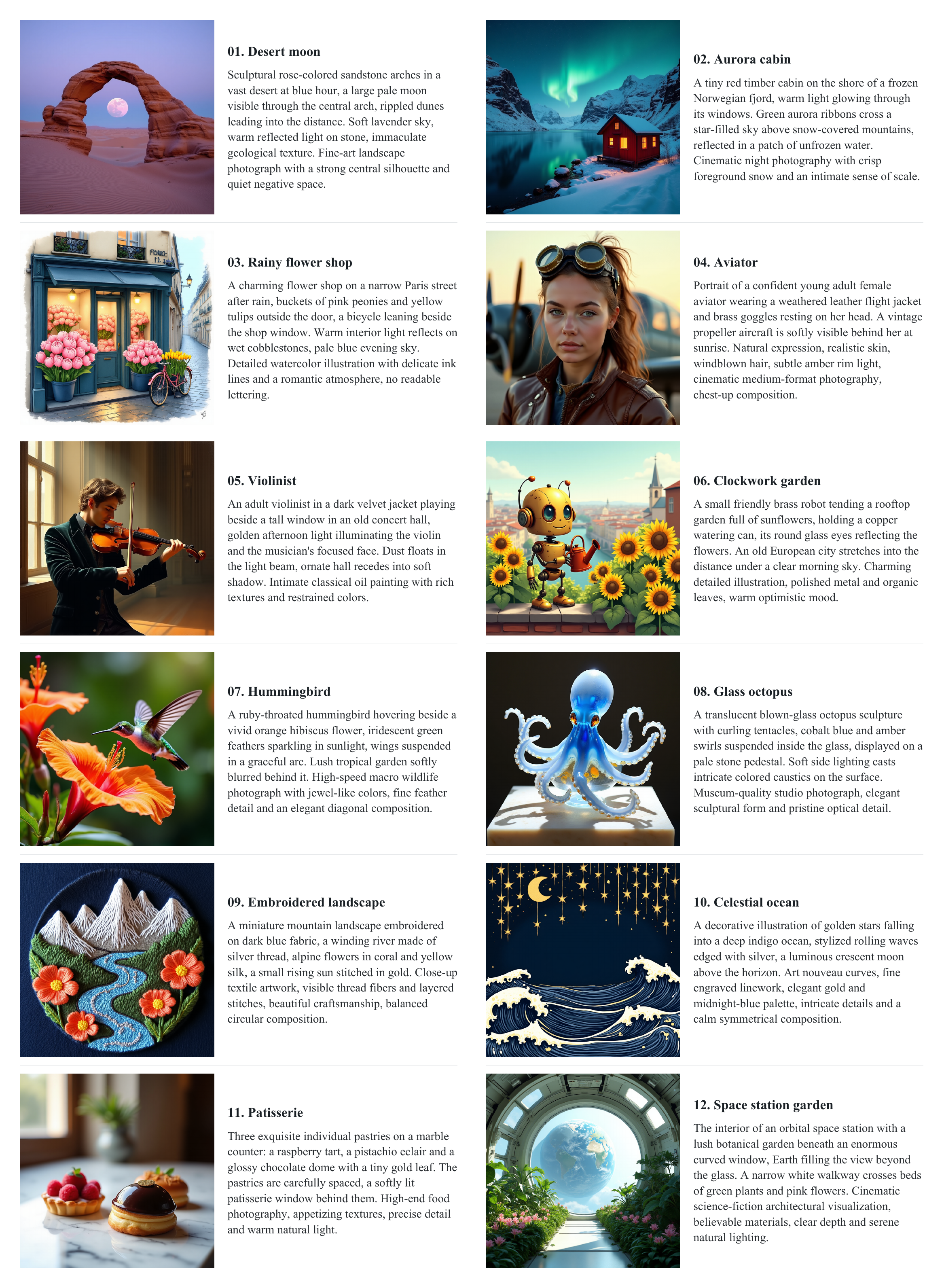}
    \caption{\textbf{Image generation samples produced by our UniCache optimized BAGEL model.} The compression rate is 60\%. The model successfully handles complex descriptions, rendering high-quality textures and correct object placements, demonstrating that our
KV cache compression framework preserves the generative performance of the model.}
    \label{fig:generation_samples}
\end{figure}

\begin{figure}[t]
    \centering
    \includegraphics[width=\linewidth]{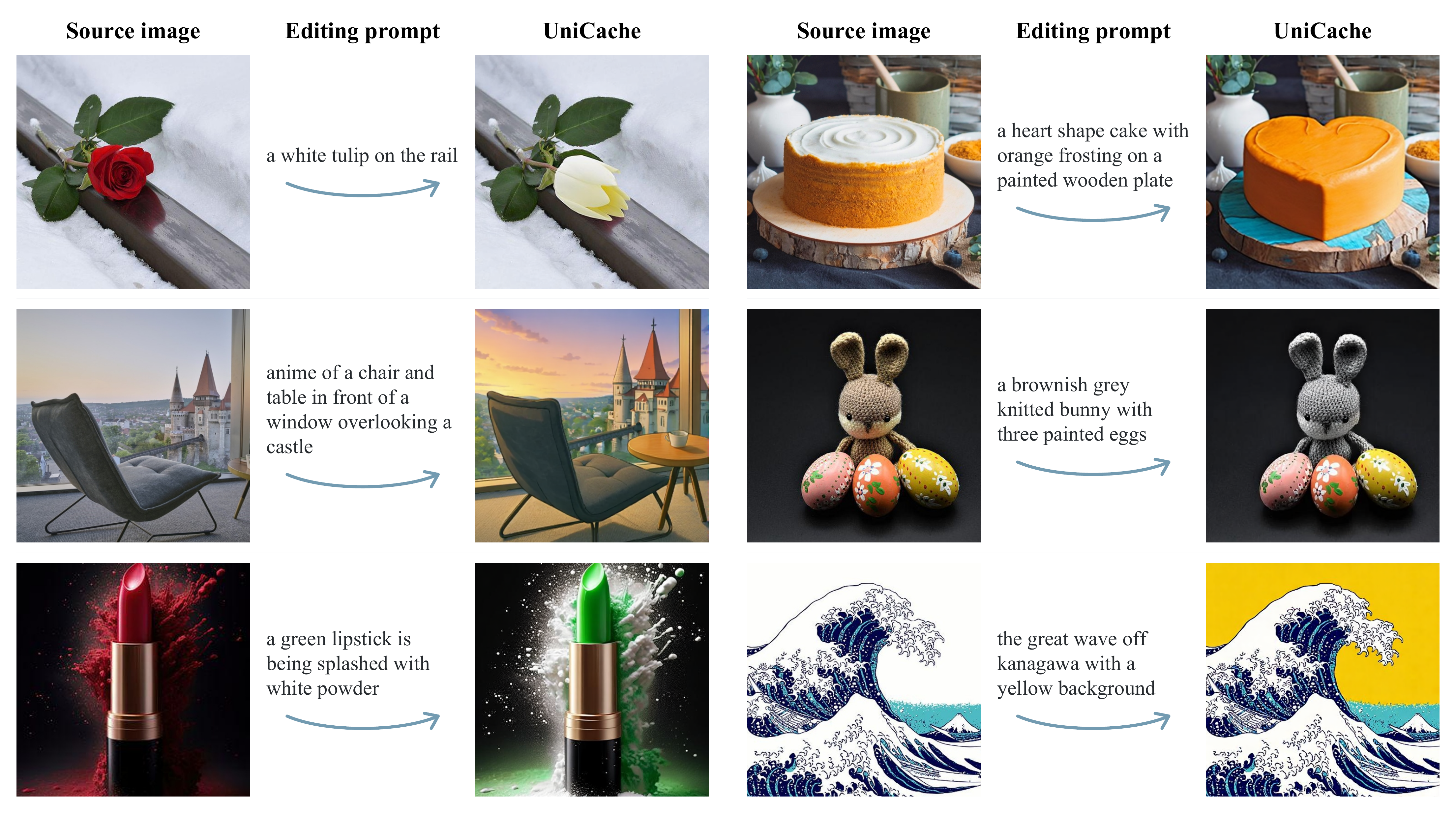}
    \caption{\textbf{Image editing samples produced by our UniCache-optimized BAGEL model.} Each group shows the source image, editing prompt, and edited output from left to right. UniCache preserves the original model's ability to follow instructions while preserving visual content unrelated to the requested edits.}
    \label{fig:editing_samples}
\end{figure}

Figures~\ref{fig:generation_samples} and~\ref{fig:editing_samples} show generation and editing samples produced by UniCache at 60\% and 80\% KV cache compression, respectively.

\section{Efficient GPU Implementation}
\label{app:open_source}

We provide both a PyTorch reference implementation and an optimized GPU inference engine for UniCache. The reference implementation uses attention masks and fake quantization to evaluate compression policies, whereas the GPU engine physically compacts or packs KV entries to reduce persistent cache storage and attention overhead.

\begin{itemize}[
leftmargin=1.2em,
labelsep=0.4em,
itemsep=1pt,
topsep=2pt,
parsep=0pt,
partopsep=0pt
]
\item \textbf{Physical, type-aware KV storage.}
For eviction-managed segments, we store only retained KV entries and their logical token indices. Quantized segments use packed low-bit representations with scales, minima, and a short BFloat16 residual tail. Protected segments retain their original representations.

\item \textbf{Batched GQA token selection.}
Attention scores are aggregated across query heads sharing the same KV head. We then perform heavy-hitter and recent-token selection independently for each KV head using batched operations, reducing Python dispatch overhead.

\item \textbf{Attention over heterogeneous cache segments.}
The engine combines cache segments under a shared softmax normalization. For multi-query denoising, we use Triton kernels to dequantize packed KVs into reusable temporary workspaces for FlashAttention. When attention is evaluated in separate chunks, their outputs are merged using log-sum-exp statistics to preserve global normalization.

\item \textbf{Efficient attention-guided allocation.}
Since FlashAttention/SDPA kernels do not explicitly materialize or store the full attention probability matrix, we implement an accelerated kernel over packed KVs that computes attention statistics at the chunk level and efficiently accumulates the attention mass required for budget allocation. Before processing each chunk, a lightweight custom kernel determines the budget allocation for the next chunk.

\item \textbf{Cached metadata and execution plans.}
At runtime, we reuse segment boundaries and type identifiers. Once a physical layout is frozen, attention bypasses repeated segment matching and generic operator preparation.

\item \textbf{Batched attention and CFG execution.}
Compatible attention calls are combined through variable-length FlashAttention, while classifier-free guidance branches are processed in configurable batches to reduce repeated kernel launches.
\end{itemize}

The engine targets BF16 inference on NVIDIA GPUs. In our efficiency evaluation, attention-guided allocation introduces only 0.45\%--1.7\% additional overhead, depending on the chunk size, which is small relative to the throughput and memory savings enabled by the overall system.

\end{document}